\pdfoutput=1

\documentclass[11pt]{article}

\usepackage[]{ACL2023}

\usepackage{times}
\usepackage{latexsym}
\usepackage{url}
\usepackage{float}
\usepackage{CJKutf8}
\usepackage{graphicx}
\usepackage{array} 
\usepackage{xcolor}

\usepackage[T1]{fontenc}

\usepackage[utf8]{inputenc}

\usepackage{microtype}

\usepackage{inconsolata}

\usepackage{ctable}

\title{Large Language Models as Annotators for Machine Translation Quality Estimation}

\author{Sidi Wang \\
 Maastricht university\\
  \texttt{sidi.wang@}\\
  \texttt{maastrichtuniversity.nl}\\\And
  Sophie Arnoult \\
 Vrije Universiteit Amsterdam\\
  \texttt{s.i.arnoult@vu.nl} \\\And
  Amir Kamran \\ 
Taus \\
  \texttt{amir@taus.net} \\} 

\begin{document}
\begin{CJK*}{UTF8}{gbsn}
\maketitle
\begin{abstract}

\end{abstract}

Large Language Models (LLMs) have demonstrated excellent performance on Machine Translation Quality Estimation (MTQE), yet their high inference costs make them impractical for direct application. 
In this work, we propose applying LLMs to generate MQM-style annotations for training a COMET model: following \citet{fernandes-etal-2023-devil}, we reckon that segment-level annotations provide a strong rationale for LLMs and are key to good segment-level QE.
We propose a simplified MQM scheme, mostly restricted to top-level categories, to guide LLM selection. We present a systematic approach for the development of a GPT-4o-based prompt, called PPbMQM (Prompt-Pattern-based-MQM). We show that the resulting annotations correlate well with human annotations and that training COMET on them leads to competitive performance on segment-level QE for Chinese-English and English-German.

\section{Introduction}
Machine Translation (MT) Quality Estimation (QE) metrics fine-tuned on human judgement data like CometKIWI \citep{rei2022cometkiwi, rei2023scaling} now perform on-par with reference-based metrics \citep{freitag-etal-2021-results,freitag-etal-2022-results,freitag-etal-2023-results,freitag-etal-2024-llms,lavie-etal-2025-findings}. 

Large Language Models (LLMs) like GPT \citep{brown-etal-2020-language} and PaLM \citep{chowdhery-etal-2024-palm,anil-etal-2023-palm} exhibit human-level performance on various benchmark NLP tasks~\citep{huang2023reasoninglargelanguagemodels}. 
\citet{kocmi2023large_gemba} were the first to utilize LLM prompting for QE, prompting GPT models to output Direct-Assessment (DA) scores. Their GPT-4-based metric GEMBA showing state-of-the-art performance at the system level, where QE metrics have to rank (test)set outputs of translation systems, and promising results at the segment level, where scores are assigned to sentence translations. 

\begin{figure}[h]
    \centering
    \includegraphics[width=0.8\linewidth]{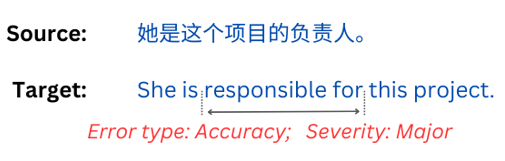}
    \caption{Example MQM annotation. Error spans are marked with error type and severity.}
    \label{fig: Intro: MQM example}
\end{figure}

\citet{fernandes-etal-2023-devil} prompt PaLM models to output Multidimensional Quality Metric (MQM) annotations \citep{burchardt2013multidimensional,lommel2014multidimensional}, which mark error spans in translations with a type and degree of severity, as shown in Figure~\ref{fig: Intro: MQM example}, and have been introduced in the WMT evaluation campaigns following \citet{freitag-etal-2021-experts}. Prompting for MQM provides interpretable results and also triggers reasoning in LLMs, and \citet{fernandes-etal-2023-devil} show that their few-shot AUTOMQM prompt has high recall on major errors and improved segment-level predictions with the larger PaLM-2 Unicorn model \citep{anil-etal-2023-palm} compared to DA-score prompting.

MQM annotations remain hard to operationalize, and other QE studies have proposed to identify error spans with their severity only and not with the error types \citep{perrella-etal-2022-matese,lu-etal-2023-toward,rei2023scaling,lu-etal-2024-error,kocmi2024errorspanannotationbalanced}, as the scores required for training QE learned metrics 
are obtained by weighing errors based on their severity, 
but also because it is sufficient for post-editing tasks \citep{kocmi2023gemba_mqm,zouhar-etal-2024-ai}. This also has resulted in the recent replacement of MQM by ESA~\cite{kocmi2024errorspanannotationbalanced} in the WMT General Machine Translation shared task~\cite{kocmi-etal-2024-findings}.
Rather than discarding all category information, we simplify the MQM scheme to top-level categories, and adding the {\em Omission} subcategory, which allows us to analyse the error types made by different LLMs.

Inference with LLMs is slow and costly \citep{rei2023scaling}, therefore we propose to use LLMs to generate {\em synthetic MQM-style annotations for finetuning learned metrics}. We make the following contributions:
\begin{itemize}
    \item We introduce a few-shot PPbMQM prompt, restricted to the top-level MQM categories for error type and the {\em omission} category.
    \item Our analysis on Chinese-English confirms that LLMs are overly critical; we show that prompting along a severity scale as a basis for score computation improves correlation with human judgments.

    \item We apply our prompt to generating training data for COMET-QE\footnote{GitHub: \url{https://github.com/Unbabel/COMET}} for Chinese-English and English-German, showing positive results with Pearson correlation compared to human annotations.

\end{itemize}

\section{Prompt development for MQM} \label{section 2}
\subsection{Experiment design}

We develop our prompt in four steps: 
1) testing LLMs on their knowledge of QE and MQM; 2) identifying issues with a first zero-shot prompt; 3) zero-shot prompting; 4) few-shot prompting. The first three steps are presented in section~\ref{sec:zero-shot}, and few-shot prompting in section~\ref{sec:few-shot}.

The MQM annotation scheme employs a refined hierarchical error typology. To simplify the task for LLMs, we initially restricted ourselves to the top-level MQM categories {\em Accuracy}, {\em Fluency}, {\em Terminology}, {\em Style}, and {\em Locale Convention}--adding the subcategory {\em Omission} for few-shot experiments--and to two severity levels, {\em Major} and {\em Minor}. 

Following \citet{white2023prompt}, we apply the {\em persona} and the {\em reflection} patterns to elicit better informed predictions, and the {\em output automater} pattern for structured output. The rationale details can be found in Appendix \ref{Appendix: basic prompt design}.

We experimented with four large LLMs, GPT-3.5, GPT-4 Turbo, GPT-4o, and LLaMA 3 70B\footnote{We used APIs provided by OpenAI and Replicate.} on Chinese-English, selecting the best performing models after each step. 

Our dataset is from the Chinese-English and English-German Expert-Based Human Evaluations (EbHE-WMT-MT)
of the 2022 and 2023 WMT general MT task \citep{freitag-etal-2021-experts}.

\subsection{Zero-shot prompting}\label{sec:zero-shot}
We first tested the translation knowledge of the four LLMs by asking questions related to QE and MQM, and scoring answers on a scale of 1 to 5. GPT-3.5 performed the poorest, scoring 14 out of 25, compared to 19 or 20 for the other models (see Table~\ref{App: step 1 result}), and was then excluded from further analysis.

\begin{table*}[h] 
    \centering

    \begin{tabular}{>{\raggedright\arraybackslash}p{0.45\linewidth}cccc} \hline
         Question&  GPT-3.5 & GPT-4 Turbo &GPT-4o&LLaMA 3\\ \hline 
         1. Can you explain what machine translation quality estimation is in 100 words?& 
     3&  4&4& 4\\ 
 2. Could you provide an overview of the Multidimensional Quality Metrics (MQM) annotation scheme in around 100 words?& 3&  4&4&4\\
 3. What are the core error categories of MQM?& 2&  3&4&4\\
 4. How can I effectively evaluate the Chinese sentence and its English translation using the MQM annotation scheme? Provide a concise response within 100 words, please.& 3&  5&5&4\\
 5. How would you, as a language model, annotate the translation of a Chinese source sentence into English using the MQM scheme? Please provide an example.& 3&  4&3&3\\ \hline
 Total score& 14& 20& 20&19\\\hline\end{tabular}
    \caption{Exploring LLMs' knowledge of QE and MQM. In each API request, we presented the LLMs with one single question. Answers are scored on a scale of 1-5 (very poor-very good) by a single annotator.}
    \label{App: step 1 result}
\end{table*}

Secondly, we designed a zero-shot
prompt that {\em relies on LLMs' inherent knowledge of MQM}, shown in Figure~\ref{App: basic prompt figure}.
\begin{figure}[h]
\texttt{\small \textbf{[System message]} You are a professional Chinese-English translator.}\\
\texttt{\small \textbf{[Instruction message]} Your task is to identify translation errors from a pair of the source Chinese sentence and the target English sentence. Please identify up to 5 errors and assign an error type for each with a severity label from "major" and "minor", using the MQM annotation scheme. 
Please provide the output only in json format:
1. error type (value scope: accuracy, style, fluency, terminology, locale convention, other)
2. marked text (the identified error in the target sentence)
3. error span index (split the target sentence with whitespace and get the marked text start and end index)
4. severity (major or minor)
5. explanation.}
\caption{Initial zero-shot prompt. This prompt was further developed by changing the error span index instruction into `split the target sentence using NLTK tokenizer and get the marked text start and end index'.}
\label{App: basic prompt figure}
\end{figure}
Testing this prompt on 10 segments showed that all three LLMs provided clear and readable values for the required fields, but generated inconsistent indices. 
We solved this by choosing the span closest to the 
indices in the case of ties. We also modified the prompt to tokenize with the NLTK tokenizer instead of by whitespace. 

Thirdly, 
we applied the revised prompt to all three models, evaluating generated annotations against human annotations on span metrics (span, severity, and error-type F1), which are calculated following \citet{blain2023findings}, and on correlation to quality scores, which are derived from MQM annotations by weighing major/minor errors by 5 and 1, respectively, following \citet{freitag-etal-2021-experts}.

\begin{table*}
\centering
\begin{tabular}{>{\raggedright\arraybackslash}p{0.17\textwidth}>{\raggedright\arraybackslash}p{0.1\textwidth}>{\raggedright\arraybackslash}p{0.14\textwidth}>{\raggedright\arraybackslash}p{0.07\textwidth}>{\raggedright\arraybackslash}p{0.07\textwidth}>{\raggedright\arraybackslash}p{0.07\textwidth}>{\raggedright\arraybackslash}p{0.06\textwidth}>{\raggedright\arraybackslash}p{0.06\textwidth}}
\hline
Model& segments& errors ({\em major})& Span& Severity &  Type &  $\rho$&  $r$\\
\hline
{\em zero-shot} \\
GPT-4 Turbo& 692& 1237 (684)&\textbf{0.359}&0.49 & 0.24&0.383$^{**}$&0.343$^{**}$\\  
GPT-4o& 957& 3146 (1256)&0.318 &\textbf{0.51} & \textbf{0.26}&0.485$^{**}$&0.382$^{**}$\\  
LLaMA 3& 885& 2286 (1143)&0.296 &0.49 & 0.16&\textbf{0.486$^{**}$}&\textbf{0.401$^{**}$}\\ \hline
{\em few-shot} \\
 GPT-4o ($m_{1-3}$)& 909& 2428 (354)& 0.333& \textbf{0.65}& \textbf{0.37}& 0.492$^{**}$&0.376$^{**}$\\ 
 GPT-4o ($m_{3}$)& 942& 1307 (367)& \textbf{0.347}& 0.62& 0.35& \textbf{0.504$^{**}$}&\textbf{0.404$^{**}$}\\\hline
\end{tabular}
\caption{Zero-shot and few-shot span metrics 
and segment-level quality score correlation with Pearson $\rho$ and Spearman $r$ on parsable segments; In the few-shot setting, we either map errors of severity 1-3 to {\em minor} errors ($m_{1-3}$), or severity-3 errors, discarding errors of severity 1-2 ($m_{3}$).  
** indicates p-value < 0.001.}
\label{Table: step 3 and 4 result}
\end{table*}
Results are presented in Table~\ref{Table: step 3 and 4 result}. Not all models are equally capable of producing parsable annotations: only 70\% of the GPT-4 Turbo annotations were parsable, compared to around 90\% for GPT-4o and LLaMA 3. 
For these two models--although the results do not compare exactly because of the differences in the parsed segments--GPT-4o produces more annotations, which align more closely to human annotations in terms of span metrics.    
\begin{table}[h]
    \centering
    \begin{tabular}{>{\raggedright\arraybackslash}p{0.12\textwidth}
    >{\raggedright\arraybackslash}p{0.028\textwidth}>{\raggedright\arraybackslash}p{0.08\textwidth}>{\raggedright\arraybackslash}p{0.028\textwidth}
    >{\raggedright\arraybackslash}p{0.07\textwidth}}
    \hline
          &  \multicolumn{2}{c}{errors}&  \multicolumn{2}{c}{major/minor}\\ 
          & gold & PPbMQM  & gold & PPbMQM  \\ \hline
  GPT4 Turbo& 1.39& 1.79 &0.24& 1.24 \\
 GPT4o&1.59 & 3.29 &0.27& 0.66 \\  
          LLaMA 3& 1.57&  2.58 & 0.28&  1.0 \\\hline
    \end{tabular}
    \caption{Average error counts per segment and {\em major/minor} ratios in human and PPbMQM annotations.}
    \label{APP: Step 3 average error numbers and major to minor ratios}
\end{table}

\begin{table}[h]
\centering
\begin{tabular}{>{\raggedright\arraybackslash}p{0.13\textwidth}>{\raggedright\arraybackslash}p{0.09\textwidth}>{\raggedright\arraybackslash}p{0.07\textwidth}>{\raggedright\arraybackslash}p{0.07\textwidth}}
\hline
& Accuracy& Style& Fluency   \\
\hline
GPT-4 Turbo & 0.55 &0.27 & 0.28\\  
GPT-4o & 0.51 &0.24 &0.29 \\  
LLaMA 3 & 0.42& 0.10 & 0.22 \\  
\hline
\end{tabular}
\caption{Error-type F1 for frequent error types.}
\label{tab:error-type-f1}
\end{table}

All LLMs are in fact stricter than human annotators, as Table~\ref{APP: Step 3 average error numbers and major to minor ratios} shows. Analyzing errors in random samples shows that they are generally not caused by hallucinations, although GPT-4o appears to be more robust than LLaMA-3 in this respect. Some errors result from ambiguity in the Chinese source sentences. For instance, GPT 4 Turbo flagged one use of the future tense as incorrect, although the exact tense of the source sentence was unclear since tense is unmarked in Chinese. In many if not most cases, LLMs are more sensitive to subtle errors than human annotators.  For instance, in the segment pair {\em 电池在用。/ The battery is working}, GPT-4o marked {\em working} as a major accuracy error, which is correct considering that 在用 means {\em in use}; in the segment pair {\em  防骗提示：/ Anti-fraud tips:}, GPT-4 Turbo flagged a minor accuracy error, which is correct considering that a better translation might be {\em Fraud prevention tips}.

Error-typing scores are low: among the most frequent error types identified by humans ({\em Accuracy}, {\em Style}, {\em Fluency}), {\em Style} and {\em Fluency} were misclassified the most (see Table \ref{tab:error-type-f1}). 
The comma splice \textit{Fluency} error is among the most common errors. We used this to inform few-shot prompting.

While LLaMA 3 annotations correlate the best with human annotations\footnote{This may seem contradictory given the lower span scores compared to GPT-4o, but correlations are calculated based on error numbers and severity labels in each segment, regardless of the span accuracy.}, we found its API stability to be lower than that of GPT-4o, experiencing multiple interruptions and errors when handling batch requests. We therefore selected GPT-4o for few-shot prompting  and QE model training \footnote{We conducted stability analyses on GPT-4o with a basic prompt variant and different system\_fingerprint values representing the backend models. Both analyses demonstrated stable behavior (see Appendix \ref{app stability analysis A.5} for results).}.

\subsection{Few-shot prompting} \label{sec:few-shot}
At this step, we improved our prompt by adding: 
1) detailed explanations for each error type; 
2) an example of {\em Omission}, as we discovered that GPT-4o did not identify this error type in the zero-shot setting\footnote{According to the MQM typology (\url{https://themqm.org/the-mqm-typology/}, access date: June 14th, 2024), \textit{Omission} is a sub-category under \textit{Accuracy}.}; 
3) an example of comma splice \textit{Fluency} error. 
Furthermore, to emulate human annotator behavior, we devised a new method for severity classification: instead of \textit{major/minor }labels, we prompt for a severity score on a scale of \textit{1-5} and then map the score ranges 4-5 to {\em Major} and ranges 1-3 to {\em Minor}. 
The final prompt version is shown in Appendix figure \ref{APP:step4 final prompt}.

\begin{table*}[!t]
\begin{tabular}{>{\raggedright\arraybackslash}p{0.12\linewidth}
    >{\raggedright\arraybackslash}p{0.1\linewidth}
    >{\raggedright\arraybackslash}p{0.06\linewidth}
    >{\raggedright\arraybackslash}p{0.11\linewidth}
    >{\raggedright\arraybackslash}p{0.06\linewidth}
    >{\raggedright\arraybackslash}p{0.11\linewidth}
    >{\raggedright\arraybackslash}p{0.06\linewidth}
    >{\raggedright\arraybackslash}p{0.11\linewidth}}

\hline
 segments & quality  & \multicolumn{2}{c}{$\rho$}  & \multicolumn{2}{c}{$r$} & \multicolumn{2}{c}{$\tau$} \\ 

&  scores & gold & PPbMQM  & gold & PPbMQM& gold & PPbMQM\\  \hline
\multicolumn{8}{c}{English} \\ \hline
4806 & 0.0-1.0 & 0.470$^{**}$& 0.513$^{**}$ & 0.410$^{**}$& 0.381$^{**}$ & 0.290$^{**}$& 0.272$^{**}$ \\ \hline
 2592  & 0.9-1.0  & 0.019& -0.157$^{**}$ & 0.020& -0.128$^{**}$ & 0.016 & -0.085$^{**}$ \\ 
          910 & 0.8-0.9 & 0.063& -0.005 & 0.077$^{*}$& 0.016& 0.053$^{*}$& 0.011 \\ 
          1304 & $<$ 0.8 & 0.434$^{**}$& 0.523$^{**}$ & 0.425$^{**}$& 0.526$^{**}$& 0.294$^{**}$& 0.370$^{**}$\\ \hline

\multicolumn{8}{c}{German} \\
\hline

1187& 0.0-1.0 & 0.503** & 0.447** & 0.325**& 0.188**& 0.243**& 0.137**\\ \hline
 796& 0.9-1.0  &  {\textminus0.041} & -0.135* & {\textminus0.012} & -0.19**& {\textminus0.004}& -0.131**\\ 
          145& 0.8-0.9 & 0.055& -0.034& 0.004& -0.047& 0.007& -0.032\\ 
          216& $<$ 0.8 & 0.692**& 0.708**& 0.565**& 0.606**& 0.402**& 0.439**\\ \hline
\end{tabular}
\caption{Bucket analysis for en-de: quality score correlation to gold annotations (Pearson $\rho$, Spearman $r$ and Kendall $\tau$) of COMET-QE trained on GPT-4o annotations (PPbMQM) vs human annotations (gold), overall and per bucket. 
p-values < 0.001 are marked with **, p-values < 0.05 with *.}
\label{table:qe-bucket analysis}
\end{table*}

Results are presented in Table~\ref{Table: step 3 and 4 result}. Few-shot PPbMQM leads to improved span metrics for GPT-4o\footnote{GPT-4o was then also able to identify \textit{Omission} errors.}, but not directly to improved correlation with human annotations. However, a severity scale allows us to set a threshold for minor errors.   

We tested different score-to-category mappings and obtained the best result by discarding annotations with severity scores 1-2, mapping only severity-3 scores to the {\em minor} category. We use our few-shot PPbMQM prompt with this mapping to generate annotations and derive quality scores for QE model training. 

\section{Downstream QE model training}
We annotated 20703 zh-en and 10121 en-de segments by applying PPbMQM \footnote{We replaced the two-shot examples with the en-de segments.}.
For each language pair, we trained two reference-free QE models with COMET using default settings, one using PPbMQM-generated annotations and a baseline using human MQM annotations. We then tested both models on randomly selected segments.

As Table~\ref{table:qe-bucket analysis} shows, for both languague pairs, the QE models trained on PPbMQM annotations performs on par with the models trained on human annotations, achieving higher Pearson correlation scores than the baseline models, albeit lower Spearman and Kendall $\tau$. In the lower quality segments, which represent a quarter of the data, they outperform the human-data trained models in all metrics. This might be related to low inter-annotator agreement for fine-grained linguistic tasks like MQM annotations\footnote{While \citet{freitag-etal-2021-experts} report higher inter-annotator agreement for MQM than Scalar Quality Metric (SQM), we find low agreement scores on EbHE-WMT-MT-2022; see Appendix Table~\ref{C3: tab: Annotator agreement 2022}.}
: the more consistent LLM annotations may be better exploited by COMET, and the difference can best be seen in segments with more errors.

\section{Discussion}

We have presented a stepwise prompting approach that progresses from knowledge testing of LLMs to zero-shot and few-shot prompting.
Our primary motivation in developing a new MQM-based prompt was not to outperform existing prompting methods, but to provide a systematic and comprehensive approach that integrates prompting with downstream QE-model training. We believe that our proposed scheme is also informative for quality evaluation with the ESA protocol \citep{kocmi2024errorspanannotationbalanced}.

Our approach and findings complement earlier work on MQM-related annotations for Quality Estimation: we redefined key error categories to balance informativeness and simplicity, keeping a subset of the MQM typology; we found, like \citet{fernandes-etal-2023-devil} and \citet{zouhar-etal-2024-ai}, that LLMs tend to overpredict errors, and we introduced a severity scaling method to effectively control their number; we did not experiment with the number of few-shot examples \citep{fernandes-etal-2023-devil,rei2023scaling}, but included example types that were frequent or undetected in the zero-shot setting.

Our prompting method generates annotations that correlate well with human annotations and that can serve for training learned QE metrics. While experimenting with more language pairs is necessary, this opens up the way to training QE models for language pairs that do not yet benefit from MQM reference annotations.
As Chinese-English and English-German are high-resource language pairs, it would be interesting, in particular, to see how LLMs perform when prompted for other lower-resource language pairs.

The COMET model trained on our synthetic data achieved better correlation with the reference annotations in the lower-quality segments. Although this was not the focus of this work, this result is particularly relevant for 
automated post-editing \citep{zerva2024findings}.

\section{Conclusion}
We have presented an MQM-based prompt for generating synthetic data for QE learned-metrics training. We keep a subset of the MQM error types and address the sensitivity of LLMs to translation errors with a severity scale, discarding low-severity errors.  
Our synthetic annotations lead to competitive performance for COMET on both Chinese-English and English-German. In future work, we plan to extend this approach to more language pairs.

\newpage
\section*{Limitations}

Our test and development datasets may have been included in the training data of LLMs, causing data leakage issues. 

The training data used for our downstream QE model training experiment only covered two language pairs on news, e-commerce, conversation, and social media domains, thus it might fail to handle tasks from other domains or/and language pairs.

Our COMET-framework-based QE models were trained with only one initialization. Future efforts should be made to test on models with multiple initializations to obtain a sufficient sample for the statistical test, strengthening the experimental evidence \citep{ulmer2022experimental_standard_nlp}.

\section*{Ethics Statement}
We used proprietary LLMs for which the source of the training data is unknown. 

\bibliography{anthology}
\bibliographystyle{acl_natbib}

\newpage
\onecolumn
\appendix
\section{Appendix}

\subsection{Basic prompt design} 
\label{Appendix: basic prompt design}
Three prompt patterns from \citet{white2023prompt} are used for our basic prompt design: \\\\
1) The persona pattern: \textit{You are a professional Chinese-English translator.} \\
This statement asks LLM to act as a specific persona -- in this case, a Chinese-English translator for our task, which should evoke traits associated with the persona.\\\\
2) The output automater pattern:\textit{ Please output a json file with the following keys; Please only generate in json format}\\
These two statements ask the LLM to provide the output in JSON format, constraining the structure of the output. If the LLM follows the instructions and produces the JSON pattern output, the manual editing effort can be reduced for automatic evaluation. Additional instructions are also given for each key to obtain output with a standard vocabulary for MQM annotations.\\\\
3) The reflection pattern: \textit{Please provide an explanation for the generated output}\\
Asking LLMs to produce an explanation for each identified error can help us understand the reasoning and assumption behind it. 
\\\\
Furthermore, the phrase {\em using the MQM annotation scheme} is added to elicit more domain knowledge that could possibly be acquired during the training phase.

\newpage
\subsection{Few-shot prompt}
\begin{figure*}[h]
\texttt{[\textbf{System message}] You are a professional Chinese-English translator.}\\
\\\texttt{[\textbf{Instruction message}] You are a professional Chinese-English translator.
Your task is to identify translation errors from a pair of the source Chinese sentence and the target English sentence. Please identify a maximum of 5 errors, assigning an error type for each with {\color{red} a severity scale from 1 (least severe) to 5 (most severe)} using the MQM annotation scheme. Please consider the following criteria for identifying errors:
\\\\
{\color{red}Accuracy: when the target translation does not accurately represent the source.\\
Omission: when the target translation is missing content present in the source text. Identify if any information has been omitted.\\
Fluency: issues with punctuation, spelling, grammar, register, inconsistency.\\
Style: when the translation is grammatically correct but uses unnatural or awkward language.\\
Terminology: inappropriate or inconsistent use of terms.\\
Locale convention: issues with formatting.}\\
\\
Please provide the output in JSON format including the following keys for each error: [“error type” (value: accuracy, omission, style, fluency, terminology, locale convention, other), “error span index” (start and end index of the marked text in the target sentence), “marked text” (the identified error in the target sentence), “severity”: {\color{red}(1-5)}]
\\
\noindent
{\color{red} Source: 产品特价的时候购买,价格不低,看评论也是很不错的产品。\\
Target: I bought it when the product was on sale, the price is not low, and it is also a very good product after reading the reviews.\\
Error 1: error type: fluency, severity: 2, marked text:, , error span index: \{start: 9, end: 9\}\\
\\
Source: 用了很久,除了低音出不来,总体还不错。\\
Target: I tried to use it for a long time, but all sounded good except for the bass.\\
Error 1: error type: accuracy, severity: 4, marked text: tried to, error span index: \{start: 1, end: 2\}\\
Error 2: error type: omission; severity: 4, marked text: 出不来, error span index: \{start: 10, end: 10\}\\
}}
\caption{The final few-shot prompt improved upon previous steps. Changes with regard to the zero-shot prompt are marked in red: severity scale; a detailed explanation of error categories; an example of \textit{Fluency}; and an example and value instruction of \textit{Omission} (since the `marked text' should come from the source sentence). Here for figure readability, we removed some quotation marks in the pseudo-JSON strings.}
\label{APP:step4 final prompt}
\end{figure*}

\newpage
\subsection{Stability analysis}\label{app stability analysis A.5}

\begin{table*}[h]
\centering
\begin{tabular}{>{\raggedright\arraybackslash}p{0.20\textwidth}>{\raggedright\arraybackslash}p{0.1\textwidth}
>{\raggedright\arraybackslash}p{0.1\textwidth}
>{\raggedright\arraybackslash}p{0.1\textwidth}
>{\raggedright\arraybackslash}p{0.1\textwidth}
>{\raggedright\arraybackslash}p{0.1\textwidth}
>{\raggedright\arraybackslash}p{0.1\textwidth}}
\hline
&segments& Span& Severity &  Type &  $\rho$&  $r$\\
\hline
basic prompt & 380& 0.307&0.46 & 0.23&0.452*&0.345*\\  
prompt variant& 386&0.308 &0.49 & 0.20&0.457*&0.376*\\  
\hline
\end{tabular}
\caption{Zero-shot results of the additional stability test with 400 output segments from GPT-4o: span metrics (span F1, severity F1, and error type F1) and segment-level correlation with Pearson $\rho$ and Spearman $r$}
\label{APP: 400 stability analysis}
\end{table*}

\begin{table*}[h]
\centering
\begin{tabular}{>{\raggedright\arraybackslash}p{0.18\textwidth}>{\raggedright\arraybackslash}p{0.1\textwidth}>{\raggedright\arraybackslash}p{0.1\textwidth}>{\raggedright\arraybackslash}p{0.1\textwidth}>{\raggedright\arraybackslash}p{0.1\textwidth}>{\raggedright\arraybackslash}p{0.1\textwidth}>{\raggedright\arraybackslash}p{0.1\textwidth}>{\raggedright\arraybackslash}p{0.1\textwidth}}
\hline
\textit{system\_fingerprint} & segments& Span& Severity &  Type &  $\rho$&  $r$\\
\hline
fp\_319be4768e&                 19155&               0.276&                         0.47&                           0.29&                         0.399*&                          0.274*\\\hline
 fp\_aa87380ac5& 714& 0.289& 0.49& 0.28& 0.478*&0.239*\\\hline
\end{tabular}
\caption{Fingerprint analysis on the PPbMQM-generated QE-training dataset: span metrics (span F1, severity F1, and error type F1) and segment-level correlation with Pearson $\rho$ and Spearman $r$}
\label{C5: qe fingerprint analysis}
\end{table*}

\newpage
\subsection{MQM inter-annotator agreement}\label{APP: human annotation agreement}
We tested the correlations of quality scores between different annotators for the same segments from the EbHE-WMT-MT
2022 (zh-en) dataset.
\begin{table}[h]
    \centering
    \begin{tabular}{cccc} \hline 
        Annotators & Number of segments & $\rho$ &  $r$\\ \hline 
        rater1, rater2 & 702 & 0.287 & 0.274 \\  
        rater1, rater3 & 1049 & 0.036 & 0.063\\ 
        rater1, rater6 & 299 & 0.059 & 0.088 \\ 
        rater1, rater5 & 11 & -0.407 &  -0.337\\ 
        rater1, rater4 & 13 & -0.143 & -0.06 \\ 
        rater2, rater3 & 515 & -0.005 & -0.042\\ 
        rater2, rater6 & 280 & 0.128 & 0.07\\  
        rater2, rater4 & 22 & 0.212 & 0.062 \\  
        rater3, rater6 & 736 & 0.043 & 0.009\\  
        rater3, rater8 & 21 & -0.101 & 0.024\\ 
        rater4, rater6 & 16 & -0.22 & -0.011\\ \hline

    \end{tabular}
    \caption{Pearson $\rho$ and Spearman $r$ correlations between annotators on the same segments in EbHE-WMT-MT 2022 (zh-en) dataset}
    \label{C3: tab: Annotator agreement 2022}
\end{table}

\end{CJK*}
\end{document}